\documentclass[10pt,twocolumn,letterpaper]{article}
\usepackage[table,dvipsnames,xcdraw]{xcolor}

\usepackage[pagenumbers]{cvpr} 

\newcommand{\name}{FunREC}

\usepackage{caption}
\usepackage{pifont} 
\usepackage[percent]{overpic}

\usepackage{listings}

\usepackage{graphicx}
\usepackage{tikz}
\usetikzlibrary{shapes, calc}
\usepackage{pifont} 
\definecolor{darkred}{rgb}{0.90,0.0,0.0}

\DeclareRobustCommand{\recordbox}{%
  \begin{tikzpicture}[baseline=(r.base)]
    \node[draw=black, rounded corners=2pt, fill=white!95!red!5,
          inner xsep=2pt, inner ysep=1.5pt, line width=0.8pt] (r)
    {\textsc{REC}\,%
     \textcolor{darkred}{\raisebox{0.1ex}{\scalebox{0.9}{\ding{108}}}}%
    };
  \end{tikzpicture}%
}

\DeclareRobustCommand{\methodnamewithgraphics}{\textsc{Fun}\hspace{0.11em}\recordbox}

\newcommand{\realdatasetname}{RealFun4D}
\newcommand{\syntheticdatasetname}{OmniFun4D}
\definecolor{qualgreen}{RGB}{81,201,120}
\definecolor{qualred}{RGB}{249,65,55}

\newcommand{\myparagraph}[1]{\noindent\textbf{#1.}\;}

\usepackage{multirow}
\usepackage[normalem]{ulem}
\usepackage{comment}

\newcommand{\best}[1]{\cellcolor{ForestGreen!20}$\mathbf{{#1}}$}
\newcommand{\second}[1]{\cellcolor{LimeGreen!20}\uline{{$#1$}}}
\newcommand{\third}[1]{\cellcolor{yellow!20}$#1$}

\DeclareRobustCommand{\bestcap}[1]{\colorbox{ForestGreen!20}{\textbf{#1}}}
\DeclareRobustCommand{\secondcap}[1]{\colorbox{LimeGreen!20}{\uline{#1}}}
\DeclareRobustCommand{\thirdcap}[1]{\colorbox{yellow!20}{#1}}

\makeatletter
\renewcommand\paragraph{\@startsection{paragraph}{4}{0pt}%
  {0.25\baselineskip minus 0.1\baselineskip}
  {-0.5em}
  {\normalfont\bfseries}}
\makeatother

\definecolor{cvprblue}{rgb}{0.21,0.49,0.74}
\usepackage[pagebackref,breaklinks,colorlinks,allcolors=cvprblue]{hyperref}

\def\paperID{700} 
\def\confName{CVPR}
\def\confYear{2026}

\title{\methodnamewithgraphics{}\\
Reconstructing Functional 3D Scenes from Egocentric Interaction Videos}

\author{
Alexandros Delitzas$^{1,2}$ \qquad
Chenyangguang Zhang$^{1 \dagger}$ \qquad
Alexey Gavryushin$^{1 \dagger}$ 
\\
Tommaso Di Mario$^{1}$ \qquad 
Boyang Sun$^{1}$ \qquad
Rishabh Dabral$^{2}$ \qquad
Leonidas Guibas$^{3}$ 
\\
Christian Theobalt$^{2}$ \qquad
Marc Pollefeys$^{1,4}$ \qquad
Francis Engelmann$^{3,5}$ \qquad
Daniel Barath$^{1}$ 
\\
{\small
$^{1}$ETH Zurich \quad
$^{2}$Max Planck Institute for Informatics \quad
$^{3}$Stanford University \quad
$^{4}$Microsoft \quad
$^{5}$USI Lugano \quad
}
}
\newcommand\blfootnote[1]{%
  \begingroup
  \renewcommand\thefootnote{}\footnote{#1}%
  \addtocounter{footnote}{-1}%
  \endgroup
}

\begin{document}

\twocolumn[{%
\renewcommand\twocolumn[1][]{#1}%
\maketitle
\begin{center}
    \captionsetup{type=figure}
    \vspace{-30pt}
    \includegraphics[width=\textwidth]{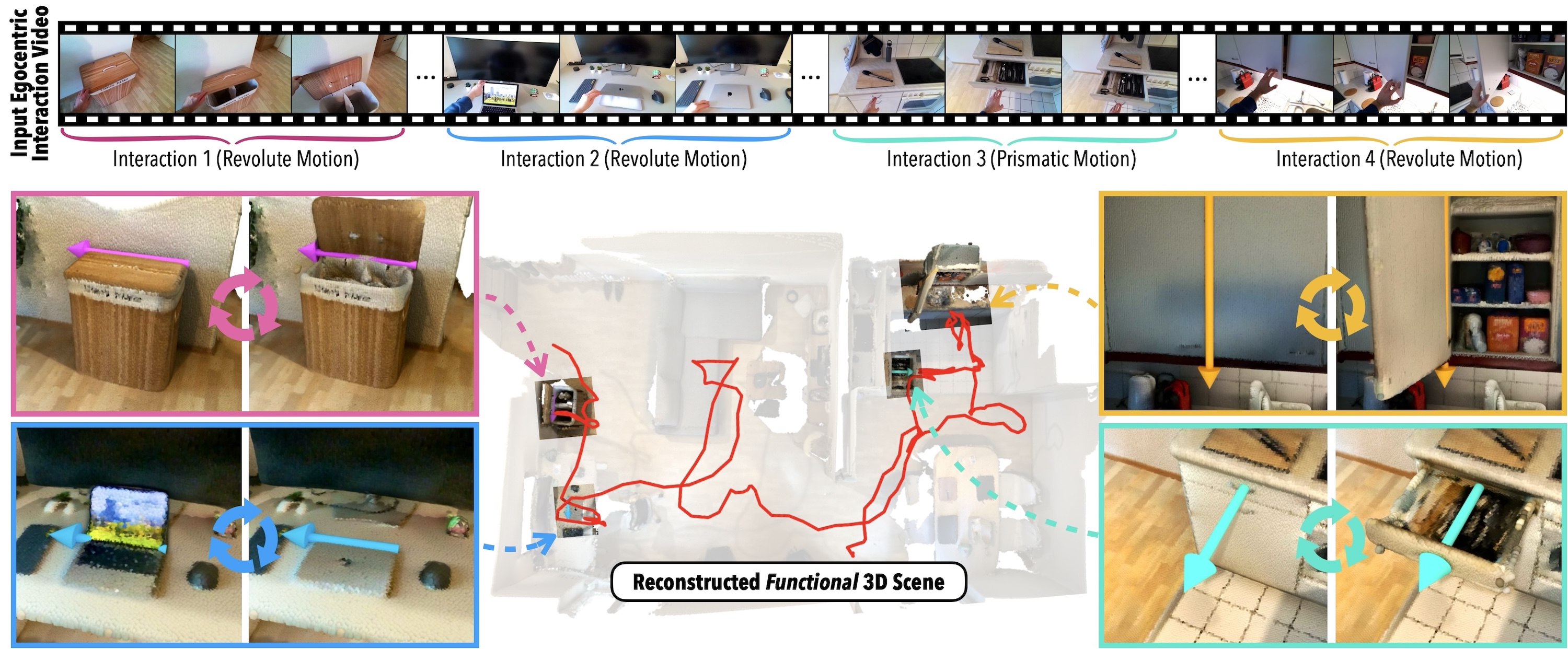}
    \vspace{-15px}
    \captionof{figure}{
    \textbf{Real-world functional digital twins.}
    \methodnamewithgraphics{} takes a single egocentric RGB-D interaction video \emph{(top)} and reconstructs a functional 3D digital twin of the environment \emph{(middle)}. The system automatically identifies articulated scene components, estimates their kinematic parameters along with per-timestep poses, and jointly reconstructs the static scene and each movable part, including interiors \emph{(see left and right)}.  The final output is a simulation-compatible 3D scene representation with fully interactable articulated elements.} 
    \label{fig:teaser}
\end{center}
\vspace{-4pt}
}]

\blfootnote{$\dagger$These authors contributed equally.}

\vspace{-10pt}
\begin{abstract}
We present \name{}, a method for reconstructing functional 3D digital twins of indoor scenes directly from egocentric RGB-D interaction videos.  
Unlike existing methods on articulated reconstruction, which rely on controlled setups, multi-state captures, or CAD priors, \name{} operates directly on in-the-wild human interaction sequences to recover interactable 3D scenes.  
It automatically discovers articulated parts, estimates their kinematic parameters, tracks their 3D motion, and reconstructs static and moving geometry in canonical space, yielding simulation-compatible meshes.  
Across new real and simulated benchmarks, \name{} surpasses prior work by a large margin, achieving up to +50\,mIoU improvement in part segmentation, 5-10$\times$ lower articulation and pose errors, and significantly higher reconstruction accuracy. 
We further demonstrate applications on URDF/USD export for simulation, hand-guided affordance mapping and robot-scene interaction. Our project page is: \href{https://functionalscenes.github.io}{functionalscenes.github.io}.
\end{abstract}
    
\vspace{-10pt}
\section{Introduction}
\label{sec:intro}

Humans make sense of the world not merely through observation, but through \emph{interacting} with it.
Reconstructing \emph{functional} 3D environments, capturing not only static geometry but also how objects move and articulate, is a central goal in computer vision, robotics, and embodied AI. 
While recent progress in 3D reconstruction and large-scale RGB-D datasets~\cite{scannet, baruch2021arkitscenes, Matterport3D17, scannetpp, Wald2019RIO, sun3d, B3DO, NYUdepthv2, sunrgbd} has advanced static scene understanding~\cite{ma2024llmsstep3dworld, scene_understanding_survey}, these datasets represent only a \emph{single state} of each environment. 
They fail to capture how scenes change under human interaction (\eg{}, doors opening, drawers sliding, fridges revealing interiors) which is essential for agents that must perceive, plan, and act in the physical world. 
This missing notion of \emph{functionality} remains a key limitation in 3D scene reconstruction.

Recent works have taken steps toward addressing this challenge, but notable limitations persist. 
MultiScan~\cite{mao2022multiscan} manually aligns multiple scans of the same room in different states (\eg, cabinet open/closed) to annotate articulated parts. 
SceneFun3D~\cite{delitzas2024scenefun3d} and Articulate3D~\cite{halacheva2024articulate3d} enrich static LiDAR scans with fine-grained functional and affordance annotations. 
Parallel work on ``digital cousins"~\cite{dai2024automated, huang2025litereality, yu2025metascenes} retrieves CAD proxies resembling static reconstructions to obtain simulated, interactive counterparts. 
While these efforts produce interactive scenes, they rely on labor-intensive multi-state captures, manual annotations, or proxy reconstruction, only weakly tied to the actual 3D geometry.
At the object level, several works~\cite{liu2023paris, jiang2022ditto, liu2025artgs, weng2024digitaltwinart, heppert2023carto, goyal2025geopard, liu2023selfsupervised, kerr2024rsrd, wu2025predictoptimized} 
recover articulated objects from demonstrations, but typically assume controlled setups, fixed cameras, or known CAD models. 
They remain limited to object-centric or synthetic scenarios, and none provides an end-to-end solution for reconstructing \emph{scene-scale, physically grounded} digital twins directly from real, egocentric interaction videos.

Our motivation is that human interaction provides the most direct and rich supervision for functional scene reconstruction. 
As people move and manipulate their surroundings, egocentric observations naturally reveal which parts articulate, around what joints, what volumes are exposed, and the associated affordances.
Building on this insight, we introduce \name{}, a system that reconstructs a coherent, articulated 3D digital twin of a real environment directly from a single egocentric RGB-D interaction video.

\name{} automatically detects articulated parts, estimates their articulation parameters, tracks scene and part motion jointly, and reconstructs both static and moving geometry, including occluded interiors. 
The result is a physically consistent, interactable 3D scene in which articulated components can be continuously manipulated along their inferred motion axes.
At its core, \name{} is a training-free, optimization-based pipeline that integrates geometric reasoning with semantic and motion priors from foundation models. It decomposes the input video into short fragments, identifies interactions via a video-language model, clusters motion trajectories into articulated components, and derives pixel-accurate interacted-part masks. We then jointly optimize part poses and articulation parameters, and fuse TSDF reconstructions using the estimated camera and part poses to obtain globally aligned functional digital twins.

We evaluate \name{} on  HOI4D~\cite{Liu_2022_hoi4d}, which is recorded in the lab showing a single object interaction, and we introduce two new egocentric interaction datasets that show a more realistic setup in real-world scenes: \realdatasetname{}, containing 351 real interaction videos from indoor spaces across 60 apartments in four countries, and \syntheticdatasetname{}, comprising 127 photorealistic sequences rendered in 12 OmniGibson scenes~\cite{li2024behavior1k, shen2021igibson}. 
Across all three datasets, \name{} outperforms state-of-the-art baselines in motion estimation, segmentation, and reconstruction quality.
Finally, we also demonstrate the practical applications of our method for URDF/USD export for simulation, affordance mapping, and robot-scene interaction.

\section{Related Work}
\label{sec:related_work}

\paragraph{Static and Interactive 3D Scene Understanding.}
Over recent years, substantial progress has been made across a range of 3D scene understanding tasks, such as segmentation~\cite{qi2017pointnet, kolodiazhnyi2024oneformer3d}, or object detection~\cite{qi2019deep, lazarow2025cubify}.
This progress has been enabled in large part by the availability of large-scale 3D datasets~\cite{scannet, baruch2021arkitscenes, Matterport3D17, scannetpp}.
While these datasets rely on \emph{static} scans, some have begun to include changes over time, \eg{}, furniture rearrangements in RIO~\cite{Wald2019RIO} or structural evolution on construction sites in NSS~\cite{sun2025nothing}.
More recently, the field has turned towards modeling \emph{articulations} and \emph{functionalities} in 3D scenes~\cite{mao2022multiscan, delitzas2024scenefun3d, halacheva2024articulate3d}.
MultiScan~\cite{mao2022multiscan} annotates articulated object parts (\eg{}, cabinet doors and drawers) by capturing each room twice, once with objects closed and once opened, and manually matching parts across states.
SceneFun3D~\cite{delitzas2024scenefun3d} and Articulate3D~\cite{halacheva2024articulate3d} focus on understanding functionality and affordances, with detailed annotations on high-resolution LiDAR scans that capture fine-grained interactive elements such as knobs and switches.
However, because these scenes are static, the causal and kinematic properties of articulated objects cannot directly be observed.
A related direction seeks to reconstruct \emph{digital cousins} of real environments~\cite{dai2024automated, huang2025litereality, yu2025metascenes} by retrieving synthetic proxies that resemble the observed scenes. 
These approaches aim to provide functional digital replicas but rely on similarity-based substitution rather than observed physical interaction.
In this work, we introduce a complementary paradigm for capturing functional, \emph{causal} 3D scene replicas. Instead of inferring articulation from static geometry, we reconstruct the scene and recover functionalities and kinematic properties directly from observed interactions.

\paragraph{3D Articulated Object Reconstruction.}

Modeling interactive scenes begins with understanding their fundamental subcomponents: the objects themselves. A substantial body of work has focused on articulated 3D objects~\cite{liu2025surveymodelinghumanmadearticulated}. Most approaches~\cite{liu2023paris, jiang2022ditto, liu2025artgs, weng2024digitaltwinart, wu2025reartgs} recover articulation by capturing multi-view RGB(-D) observations across discrete object states and leveraging these to infer articulation parameters and reconstruct part-level geometry. However, the reliance on multi-view capture makes the data acquisition complex and limits practicality.
Another line of research integrates vision and language models to infer articulation by generating executable code~\cite{zhao2024real2code} or retrieving CAD models from visual input~\cite{le2025articulate}, while others~\cite{heppert2023carto, goyal2025geopard, liu2023selfsupervised, xia2025drawer, opdmulti, jiang2022opd, huang2025react3d} operate from single observations.
Nevertheless, the generalization of these works to in-the-wild settings is hindered by inaccurate CAD model retrievals or the reliance on 2D articulation estimators trained on small-scale datasets. More recently, a new family of methods has emerged that jointly capture geometry and motion from a human demonstration video~\cite{kerr2024rsrd, wu2025predictoptimized}. However, these approaches rely on demonstrations recorded with a static camera and require pre-scanned 3D object models, which restricts their applicability in unconstrained, real-world settings. Concurrently with our work, iTACO~\cite{peng2025itaco} and VideoArtGS~\cite{liu2025videoartgs} reconstruct articulated objects from dynamic videos, but operate in object-centric setups and do not address scene-scale reconstruction in the wild. ArtiPoint~\cite{arti25werby} estimates articulation axes from interaction videos at scene level, but does not recover part geometry.

\paragraph{Tracking Interactions in Dynamic 3D Scenes.}
While several impactful datasets have been proposed for dynamic object interactions ranging from rigid~\cite{Liu_2022_hoi4d, banerjee2024hot3d, wang2025hocap, Kwon_2021_ICCV} to articulated objects~\cite{fan2023arctic, kim2024parahome}, they primarily operate in a controlled, table-top setting with objects  artificially scattered in the scene.
Consequently, their applicability to  in-the-wild scenes, \eg, real apartments, remains limited.
Recent progress in 6D object pose tracking~\cite{wen2023bundlesdf, wen2021bundletrack, foundationposewen2024} has enabled several works that track objects and scene changes from egocentric videos~\cite{LostFoundBehrens, egogaussian, guzov24ireplica} in real scenes.
Yet, they either only track rigid changes, or assume a given 3D scene scan pre-labeled with part-wise motions. 
Several recent works have focused on monocular 4D reconstruction~\cite{zhang2024monst3r,li2025megasam, yuan2024self-supervised, cut3r, som2024, lei2024mosca}.
Typically trained to reconstruct the dynamic point maps and camera trajectories in a feedforward fashion, such methods do not understand the scene and object semantics, let alone the articulation parameters of the individual objects.
Furthermore, they often operate on short video sequences and struggle with depth inconsistencies and flickering artifacts in the presence of occlusions, which are routinely encountered during in-the-wild egocentric scene interactions.
While most methods do not predict 3D motion tracks, the ones that do~\cite{xiao2025spatialtracker, st4rtrack2025}  struggle in the presence of occlusions due to interactions.
Meanwhile, 2D and 3D point trackers~\cite{karaev23cotracker, karaev24cotracker3, tapip3d, wang2023omnimotion, SpatialTracker, ICLR2025_Delta, harley2025alltracker, rajic2025mvtracker} provide useful motion priors but can be noisy. 
How to leverage them to robustly track the evolution of scene states in real-world articulated 3D environments remains an open question.
We propose a training-free approach that leverages the semantic and motion priors of foundational models to robustly \textit{discover} functionalities and track scene states from casually captured egocentric interaction videos.

\begin{figure*}[t]
    \centering
    \includegraphics[width=0.945\textwidth]{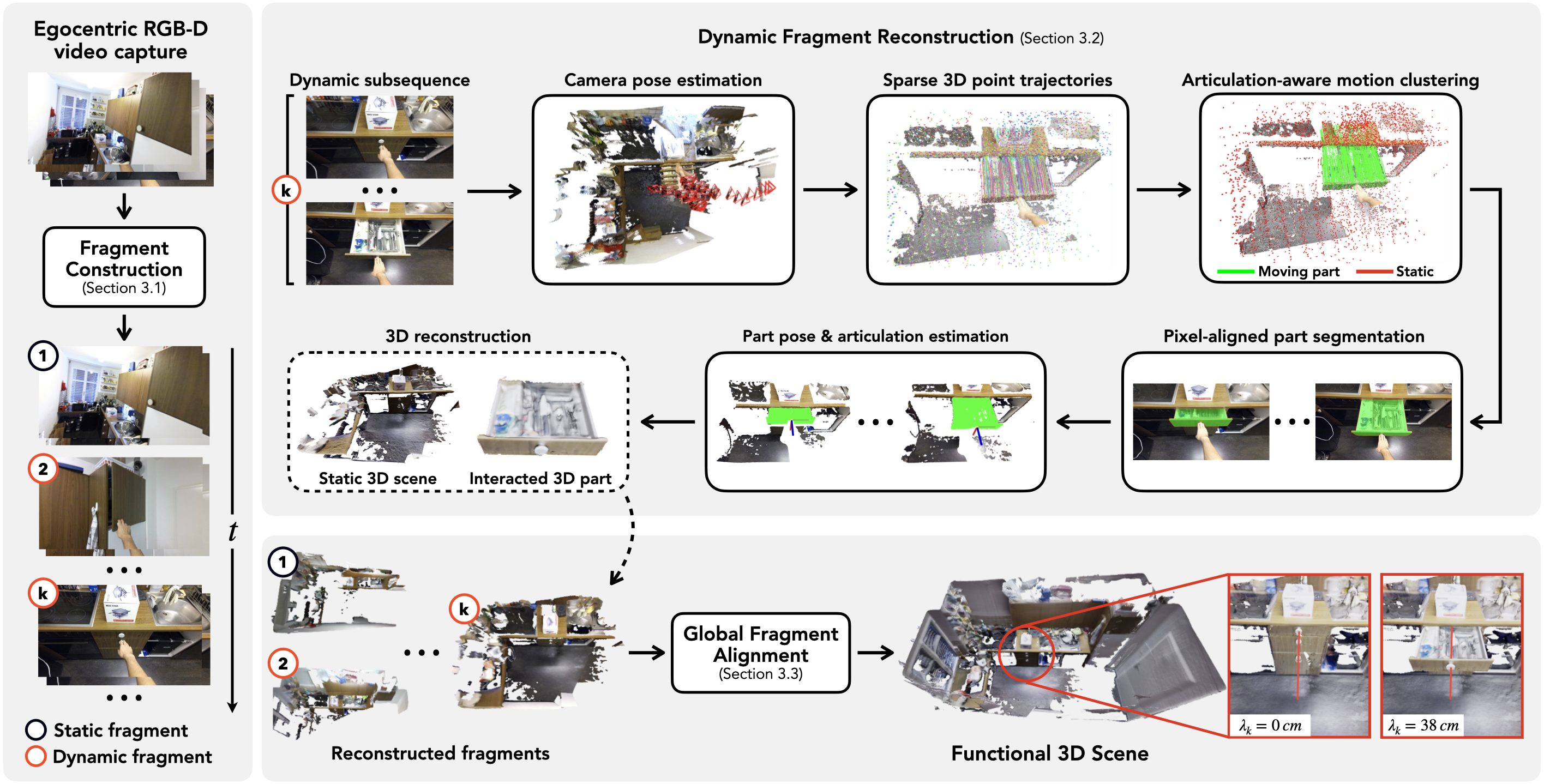}
    \vspace{-3px}
    \caption{\textbf{Method overview}. 
    Given an egocentric RGB-D interaction video, \name{} first divides it into static and dynamic fragments. 
    For each dynamic fragment, it estimates camera poses, computes sparse 3D point trajectories, and clusters them into articulated components via articulation-aware motion modeling. 
    The interacting part is then segmented to obtain dense masks and reconstructed together with the static scene. 
    Part pose and articulation parameters are optimized jointly to yield consistent motion across frames. 
    Finally, all reconstructed fragments are globally aligned to produce a coherent, functional 3D digital twin where articulated parts can be interactively manipulated.}
    \label{fig:method_overview}
\vspace{-15px}
\end{figure*}

\section{Proposed Method}
\label{sec:method}

\paragraph{Problem formulation.}
Let $\mathcal{V} = \{(I_i, D_i)\}_{i=1}^{N}$ denote an RGB-D video consisting of $N$ frames, where $I_i$ and $D_i$ represent the RGB image and the depth map at time $i$, respectively, and camera intrinsics are known.  
The scene is composed of a static background geometry $\mathcal{P}^s$ and a set of $K$ {articulated parts} $\{\mathcal{P}^{m_k}\}_{k=1}^{K}$.  
Each part $m_k$ is parameterized by a set of articulation parameters $\phi_k$ and a sequence of rigid transformations $\{T_i^{m_k}\}_{i=1}^{N}$, where each $T_i^{m_k} \in \mathrm{SE}(3)$ maps the part from its canonical coordinate frame to the world frame at time $i$.
We model two articulation types:

\noindent\textit{Prismatic joints.}
For sliding object parts such as drawers, we define a unit translation axis $\mathbf{a} \in \mathbb{S}^2$ and a scalar displacement $\lambda_i \in \mathbb{R}$ at frame $i$, where $\mathbb{S}^2$ is the unit sphere.
The transformation at frame $i$ is then given by:
\vspace{-1.5pt}
\begin{equation}
\small
    \mathcal{T}_{\text{pris}}(\mathbf{a}, \lambda_i) =
    \begin{bmatrix}
        I & \lambda_i \mathbf{a} \\
        \mathbf{0}^\top & 1
    \end{bmatrix},
    \label{eq:prismatic}
\end{equation}
where $I$ is the $3{\times}3$ identity matrix.

\noindent\textit{Revolute joints.}
For rotating object parts such as doors, we define a unit rotation axis $\mathbf{a} \in \mathbb{S}^2$, a pivot point $\mathbf{p} \in \mathbb{R}^3$ lying on the axis and closest to the scene origin (\ie, $\mathbf{a}^\top \mathbf{p}=0$), and a rotation angle $\theta_i \in \mathbb{S}^1$ at frame $i$, where $\mathbb{S}^1$ is the unit circle.  
The corresponding transformation is:
\vspace{-1.5pt}
\begin{equation}
\small
\mathcal{T}_{\text{rev}}(\mathbf{a}, \mathbf{p}, \theta_i) =
\begin{bmatrix}
R(\mathbf{a}, \theta_i) & (I - R(\mathbf{a}, \theta_i))\,\mathbf{p} \\
\mathbf{0}^\top & 1
\end{bmatrix},
\label{eq:revolute}
\end{equation}
where $R(\mathbf{a}, \theta)$ is the $3{\times}3$ rotation matrix around axis $\mathbf{a}$ by angle $\theta$.

Thus, the pose of part $m_k$ at time $i$ is:
\vspace{-1.5pt}
\[
\small
T_i^{m_k} =
\begin{cases}
\mathcal{T}_{\text{pris}}(\mathbf{a}_k, \lambda_{i,k}), & \text{if } m_k \text{ is prismatic},\\[4pt]
\mathcal{T}_{\text{rev}}(\mathbf{a}_k, \mathbf{p}_k, \theta_{i,k}), & \text{if } m_k \text{ is revolute.}
\end{cases}
\]

\subsection{Fragment Construction}
The input video $\mathcal{V}$ is divided into temporally contiguous fragments $\mathcal{V}_k$ (\cref{fig:method_overview}), such that $\mathcal{V} = \bigcup_{k=1}^{K} \mathcal{V}_k$.  
Each fragment is automatically classified as either \emph{static} (no interaction) or \emph{dynamic} (interaction with an articulated part) using a video-language model (VLM)~\cite{comanici2025gemini25pushingfrontier}.  
For each dynamic fragment, the VLM also predicts the articulation type $\sigma_k \in \{\text{prismatic}, \text{revolute}\}$.  
Each fragment is processed independently as described below.

\subsection{Dynamic Fragment Reconstruction}

\paragraph{Camera pose estimation.}
Dense point correspondences are extracted between consecutive RGB-D frames using RoMA~\cite{edstedt2024roma}.  
We filter these matches using per-frame hand and interacted-object masks from VISOR~\cite{VISOR2022}: points whose pixels lie inside the hand mask $\{M_i^h\}$ are discarded, and those inside the interacted-object mask $\{M_i^{obj}\}$ are down-weighted by scaling their RoMA confidence scores. 
Each surviving 2D correspondence is lifted into 3D, and the relative camera motion between frames is estimated using SupeRANSAC~\cite{barath2025superansacransacrule}.  
A fragment-level pose graph optimization ensures globally consistent camera poses $\{T_i^c\}$.

\paragraph{Sparse 3D trajectories.}
We obtain sparse 3D point trajectories directly in the world frame using TAPIP3D~\cite{tapip3d}.  
This produces track positions $\tau \in \mathbb{R}^{T \times N \times 3}$ and per-frame visibility scores $o \in \mathbb{R}^{T \times N}$, where $T$ is the number of tracked points.  
New tracks are periodically initialized on a uniform 2D grid to capture surfaces revealed during interaction.
Having established sparse 3D trajectories, next we assign tracks to the interacted articulated part. 

\paragraph{Articulation-aware motion clustering.}
Let $\tau_l = \{\tau_{l,i}\}_{i=1}^{N}$ denote the 3D coordinates of the $l$-th track.  
Tracks with negligible motion are discarded by thresholding their maximal displacement with $\epsilon_s$.  
For each remaining track, we estimate a per-track articulation hypothesis $\hat{\phi}_l$ consistent with the fragment-wide articulation type $\sigma_k$ as:
\[
\small
\hat{\phi}_l =
\begin{cases}
(\hat{\mathbf{a}}_l, \{\hat{\lambda}_{l,i}\}), & \text{if } \sigma_k = \text{prismatic},\\[3pt]
(\hat{\mathbf{a}}_l, \hat{\mathbf{p}}_l, \{\hat{\theta}_{l,i}\}), & \text{if } \sigma_k = \text{revolute}.
\end{cases}
\]
The motion of each track is fitted by a 3D line (prismatic) or circle (revolute).  
We compute predicted track positions $\hat{\tau}_{l,i}$ by transforming the initial point $\tau_{l,1}$ with the estimated motion parameters as follows:
\[
\small
\hat{\tau}_{l,i} =
\begin{cases}
\mathcal{T}_{\text{pris}}(\hat{\mathbf{a}}_l, \hat{\lambda}_{l,i}) \tau_{l,1}, & \text{if prismatic},\\[4pt]
\mathcal{T}_{\text{rev}}(\hat{\mathbf{a}}_l, \hat{\mathbf{p}}_l, \hat{\theta}_{l,i}) \tau_{l,1}, & \text{if revolute.}
\end{cases}
\]
A track is retained if its average fitting error satisfies
%
\[
\small
\frac{1}{|\tau_l|} \sum_{\tau_{l,i} \in \tau_l} \|\tau_{l,i} - \hat{\tau}_{l,i}\|_2 < \epsilon_f.
\]
Tracks passing this filter are clustered using HDBSCAN~\cite{hdbscan} according to the similarity of their fitted joint parameters (axis, pivot, and motion pattern), forming motion clusters that represent independently moving parts.

To identify the cluster corresponding to the manipulated part, we compare clusters against 2D interaction evidence.  
Let $\mathcal{C}_i^{obj}$ be the confidence associated with the interacted-object mask $\mathcal{M}_i^{obj}$ at frame $i$.  
For each cluster $\gamma$, we compute a consistency score as:
\vspace{-5pt}
\begin{equation}
\small
    s_\gamma = \sum_{l \in \gamma} \sum_i o_{l,i} \cdot \mathcal{C}_i^{obj} \cdot \mathbb{I}\!\left[\pi(\tau_{l,i}) \in \mathcal{M}_i^{obj}\right],
    \label{eq:cluster_score_consistent}
\end{equation}
where $\pi(\cdot)$ denotes projection to the image plane.  
The cluster with the highest score $s_\gamma$ is selected as the interacted part, yielding the moving track set $\tau^m$ and the static set $\tau^s$.

\paragraph{Pixel-aligned part segmentation.}
Given the moving track set $\tau^m$, our goal is to obtain a dense, pixel-aligned segmentation mask of the articulated part for each frame. Since $\tau^m$ is sparse and may contain outliers, directly projecting tracks or prompting a segmentation model is unreliable. To obtain a robust segmentation, we combine geometric evidence from the tracks with image-level semantic grouping. 

We first select a set of keyframes $\{I_q\}_{q=1}^{Q}$ uniformly across the fragment. On each keyframe $I_q$, we apply SAM’s automatic mask generator to produce an over-segmentation
    $G_q : \{1,\dots,H\} \times \{1,\dots,W\} \rightarrow \mathbb{N}$,
where $G_q(p)$ assigns each pixel $p$ to a semantic region (mask) hypothesis, and $H, W$ denote the image height and width. The static and moving 3D tracks are projected into the keyframe using the corresponding camera pose $T_q^{c}$ as $\pi(\tau_q^m)$ and $\pi(\tau_q^s)$, where $\pi : \mathbb{R}^3 \rightarrow \mathbb{R}^2$ is the projection function. 

For each semantic region $r$ in $G_q$, we count the number of projected moving and static tracks contained within it:
\begin{equation}
\small
n_r^m = \sum_{p \in r} \mathbb{I}\!\left[p \in \pi(\tau_q^m)\right], \quad
n_r^s = \sum_{p \in r} \mathbb{I}\!\left[p \in \pi(\tau_q^s)\right],
\end{equation}
where $\mathbb{I}$ is an indicator. 
We compute a motion ratio, quantifying if the current region contains the moving part, as
$\gamma_r = n_r^m / (n_r^m + n_r^s + \epsilon)$,
where $\epsilon$ prevents division by 0. Regions with $\gamma_r > \eta_m$ are classified as belonging to the moving articulated part, where $\eta_m$ is a threshold. This produces a motion mask $\mathcal{M}_q^{sm}$ on each selected keyframe $q$ as:
\begin{equation}
\small
    \mathcal{M}_q^{sm}(p) =
    \begin{cases}
        1, & \text{if } G_q(p) \in \{r \mid \gamma_r > \eta_m\},\\
        0, & \text{otherwise.}
    \end{cases}
\end{equation}
The keyframe motion masks $\{\mathcal{M}_q^{sm}\}$ serve as prompts for SAM2's video propagation module~\cite{ravi2024sam2}, producing a temporally consistent sequence of articulated
part masks $\{\mathcal{M}_i^{m}\}_{i=1}^{N}$.
These masks accurately delineate the articulated part across the entire fragment, providing the pixel-level support necessary for dense part reconstruction in subsequent steps.

\paragraph{Part pose and articulation estimation.}
Given the moving track set $\tau^m = \{\tau_{l,i}^m\}$ and corresponding visibility scores $o^m = \{o_{l,i}^m\}$, our goal is to recover the globally consistent part poses $\{T_i^{m}\}_{i=1}^{N}$ and articulation parameters $\phi^m$ describing the motion of the part across all frames. 
Each pose $T_i^{m} \in \mathrm{SE}(3)$ maps the canonical part coordinate frame to the world frame at time $i$. The articulation parameters $\phi^m$ encode the joint model, defined as $(\mathbf{a}, \{\lambda_i\})$ for prismatic motion or $(\mathbf{a}, \mathbf{p}, \{\theta_i\})$ for revolute motion.

For each pair of frames $(i,j)$ within a fragment, we construct 3D-3D correspondences between visible points in $\tau^m$ and estimate the relative transformation of the part, denoted $T_{i \rightarrow j}^{m} \in \mathrm{SE}(3)$, using SupeRANSAC~\cite{barath2025superansacransacrule}. Each correspondence is weighted by the product of its visibility scores $o_{l,i}^m \cdot o_{l,j}^m$. The resulting set of relative transforms forms a pose graph connecting all part poses $\{T_i^{m}\}$ in the fragment.

To jointly recover the absolute part poses and articulation parameters, we minimize the following objective:
\begin{align}
\small
\mathcal{L}(T^m, L^m, \phi^m) =
\sum_i f(T_i^m, T_{i+1}^m, T_{i \rightarrow i+1}^m) \nonumber \\
+ \sum_{i,j} l_{ij}^m \, f(T_i^m, T_j^m, T_{i \rightarrow j}^m) 
+ \mu \sum_{i,j} (\sqrt{l_{ij}^m} - 1 )^2,
\label{eq:part_pose_graph}
\end{align}
where $l_{ij}^m \in [0, 1]$ are optimized loop-closure confidences, and $\mu$ controls their regularization. The term 
\begin{align*}
\small
f(T_i^m, T_j^m, T_{i \rightarrow j}^m) & = e_{ij}^\top \Omega_{ij} e_{ij},
\end{align*}
where $e_{ij} = \log_{\mathrm{SE}(3)}\left((T_i^m)^{-1} T_j^m T_{i \rightarrow j}^m\right)$
measures the discrepancy of the estimated relative transformation $(T_i^m)^{-1} T_j^m$ and the observed transformation $T_{i \rightarrow j}^m$, weighted by the information matrix $\Omega_{ij}$. The logarithm $\log_{\mathrm{SE}(3)}(\cdot)$ maps the residual to the tangent space of $\mathrm{SE}(3)$. 

The articulation parameters $\phi^m$ are initialized via least-squares fitting to the observed tracks $\tau^m$, providing an initial estimate of the joint axis and motion states. Both $\{T_i^m\}$ and $\phi^m$ are then jointly refined through non-linear optimization using Ceres Solver, employing manifold optimization to ensure the articulation parameters remain on their respective manifolds as defined in our problem formulation~\cref{sec:method}.

\paragraph{Reconstruction and interactive manipulation.}
Given the per-pixel segmentation masks and estimated poses, we reconstruct the geometry using two separate truncated signed distance function (TSDF) volumes: one for the static background and one for the articulated part. The static TSDF volume is integrated in the world coordinate frame using the estimated camera poses $\{T_i^c\}$, while excluding dynamic regions corresponding to hand and moving-part pixels, as indicated by binary masks $\mathcal{M}_i^h$ and $\mathcal{M}_i^{m}$, respectively.
The articulated part is reconstructed in its canonical coordinate frame. For each frame $i$, the depth map is first transformed from the camera frame into the part frame using $(T_i^m)^{-1} T_i^c$, and only pixels belonging to the moving-part mask $\mathcal{M}_i^{m}$ are fused. This yields a clean canonical 3D model $\mathcal{P}^m$, free from camera motion and occlusions.
After both TSDF volumes are fused, meshes are extracted for the static scene $\mathcal{P}^s$ and articulated part $\mathcal{P}^m$. The complete scene at time $i$ is obtained as:
    $\mathcal{P}_i = \mathcal{P}^s \cup T_i^m(\mathcal{P}^m)$, 
where $T_i^m(\mathcal{P}^m)$ applies the estimated pose of the part to its canonical geometry, placing it in the global frame. 
To support interactive manipulation, we estimate the feasible motion range of the articulation parameters from the tracked motion trajectory:
    $\lambda \in [\lambda_{\min}, \lambda_{\max}]$ 
    or $\theta \in [\theta_{\min}, \theta_{\max}]$,
allowing the reconstructed scene to be rendered or physically simulated at any intermediate state consistent with the articulation $\phi^m$.

\subsection{Global Fragment Alignment}
%
Each dynamic fragment yields a local submap $\mathcal{S}_k$ containing its reconstructed static geometry and articulated parts. To produce static fragment submaps, we perform only camera pose estimation, since the scene remains static. To form a globally consistent scene, we align all submaps $\{\mathcal{S}_k\}_{k=1}^{K}$ within a shared coordinate frame. For each submap pair $(\mathcal{S}_k, \mathcal{S}_{k'})$, geometric correspondences are extracted, and a relative rigid transformation is estimated using PREDATOR~\cite{predator}. Loop closures are accepted only if the point alignment root-mean-square error (RMSE) falls below a predefined threshold.
These relative transformations define a scene-level pose graph over the submaps, which we optimize to obtain globally consistent submap poses. The aligned submaps are finally fused into a unified TSDF volume, from which we extract the complete static scene $\mathcal{P}^s$ and the set of all articulated part meshes $\{\mathcal{P}^{m_k}\}$.

\begin{figure*}[ht]
    \centering
    \includegraphics[width=0.95\textwidth]{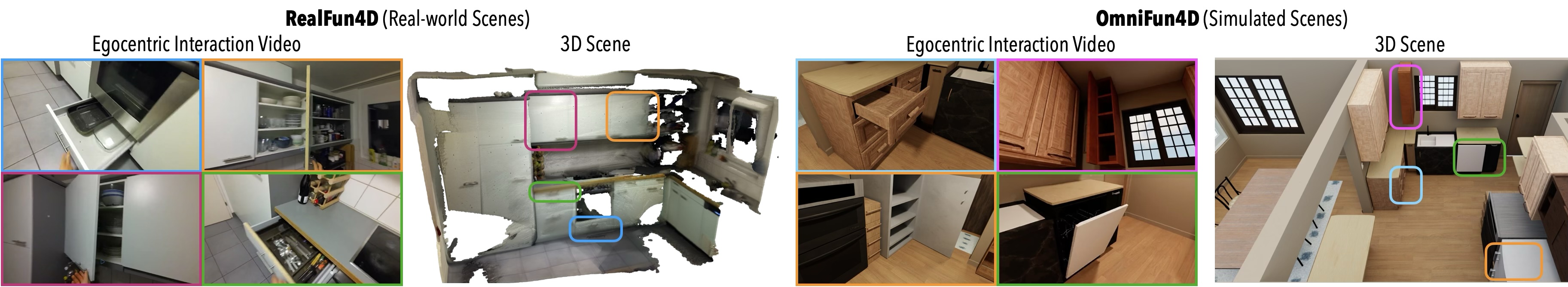}
    \caption{\textbf{Proposed Datasets.} We introduce two datasets for functional 3D scene reconstruction and evaluation: RealFun4D \emph{(left)}, capturing egocentric interactions in real scenes, and OmniFun4D \emph{(right)}, providing photorealistic simulated interactions in synthetic scenes.}
    \label{fig:datasets}
\end{figure*}

\section{Data Collection}
\label{sec:data_collection}

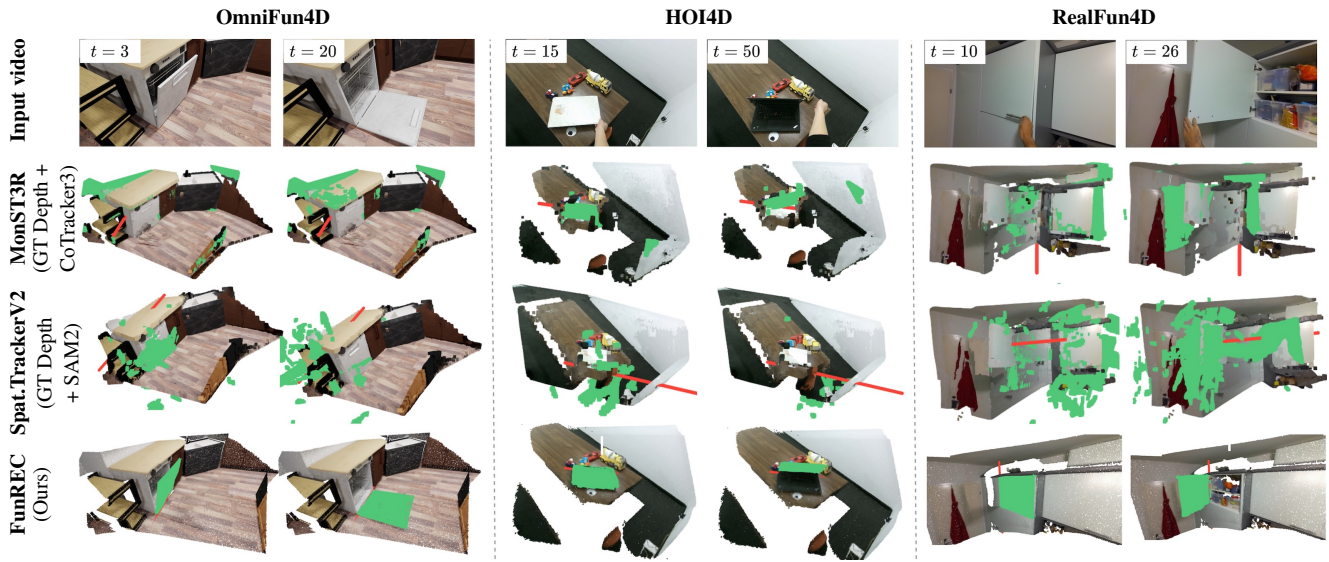
\begin{figure*}[ht]
\vspace{20px}
\begin{flushright}
\begin{overpic}[width=0.95\textwidth]{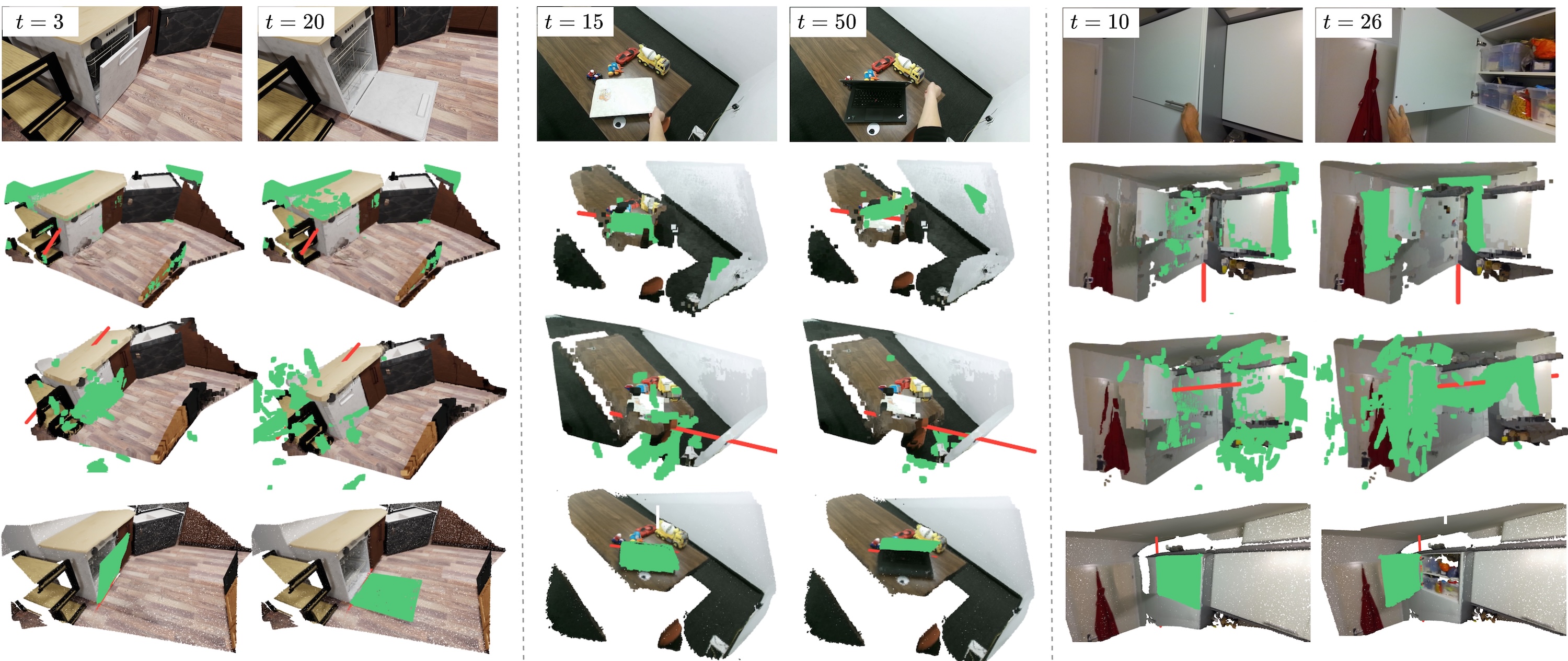}
  \put(11, 43){\footnotesize \textbf{OmniFun4D} } 
  \put(47, 43){\footnotesize \textbf{HOI4D}}
  \put(78, 43){\footnotesize \textbf{RealFun4D} }

  \put(-5.5, 33.5){\rotatebox{90}{\footnotesize \shortstack{\textbf{Input video}}}}
  \put(-5.5, 23){\rotatebox{90}{\footnotesize \shortstack{\textbf{MonST3R}\\(GT Depth +\\CoTracker3)}}}
  \put(-5.5, 10){\rotatebox{90}{\footnotesize \shortstack{\textbf{Spat.TrackerV2}\\(GT Depth\\+ SAM2)}}}
  \put(-5.5, 2){\rotatebox{90}{\footnotesize \shortstack{\textbf{\name{}}\\(Ours)}}}
\end{overpic}
      \end{flushright}
\vspace{-8px}
    \caption{\textbf{Qualitative comparisons}. We show qualitative comparisons between baselines and our method. For each method, we accumulate the reconstructed point clouds of both the articulated part and the static scene across all timesteps, and visualize them under two selected scene states. \textcolor{qualgreen}{Green} indicates the articulated part, and \textcolor{qualred}{red} lines denote the estimated articulation axes.
    }
    \label{fig:qualitative_results}
\vspace{-6px}
\end{figure*}

\vspace{-3px}
HOI4D~\cite{Liu_2022_hoi4d} is the only prior dataset for our task but offers low-motion, single-object interactions and no full scenes. Thus, we collect two new egocentric datasets with realistic, diverse interactions in real and simulated scenes (\cref{fig:datasets}).

\paragraph{Real-world scenes: \realdatasetname{}.}
Our real dataset contains 351 \emph{in-the-wild} human-scene interactions recorded across 60 apartments in four countries. Each sequence is captured with a head-mounted Azure Kinect DK ($1920\times1080$, 15 FPS), providing synchronized RGB and depth. We annotate interaction intervals and textual descriptions, label per-frame 2D hand and part masks for dynamic-region filtering and static-scene reconstruction~\cite{Liu_2022_hoi4d}, and mark articulation joints along with per-frame 2D part tracks to enable alignment and full 3D part reconstruction.

\paragraph{Simulated scenes: \syntheticdatasetname{}.}
We further record 127 interactions in 12 OmniGibson~\cite{li2024behavior1k} scenes derived from iGibson~\cite{shen2021igibson}. A human operator navigates the environment and triggers scripted interactions; camera poses and events are logged and replayed offline to render high-quality RGB-D and masks using NVIDIA RTX Path Tracing~\cite{RTXPT}. We add stochastic Gaussian perturbations to the camera poses to emulate natural head motion. Further details are provided in the supplementary material.

\vspace{-2px}
\section{Experiments}
\label{sec:experiments}

\begin{table*}[t]
\centering
\resizebox{\linewidth}{!}{
    \begin{tabular}{l*{12}{c}}
        \toprule
        \multirow{2}{*}{\textbf{Methods}} & \multicolumn{4}{c}{\textbf{\syntheticdatasetname{}}} & \multicolumn{4}{c}{\textbf{HOI4D}} & \multicolumn{4}{c}{\textbf{\realdatasetname{}}} \\
        \cmidrule(lr){2-5} \cmidrule(lr){6-9} \cmidrule(lr){10-13} 
         & {Axis ($^\circ$)} & {Pos (m)} & {State ($^\circ$/m)} & {Fail (\%)} & {Axis ($^\circ$)} & {Pos (m)} & {State ($^\circ$/m)} & {Fail (\%)} & {Axis ($^\circ$)} & {Pos (m)} & {State ($^\circ$/m)} & {Fail (\%)} \\
        \midrule
        MonST3R~\cite{zhang2024monst3r} (ICP) 
        & $52.7 / 61.5$ & $1.14$ & $66.2 / 0.27$ & \second{11.7} 
        & $67.3 / 66.1$ & $0.45$ & $60.8 / 0.06$ & \best{0.0} 
        & $54.6 / 59.1$ & $0.68$ & $65.9 / 0.21$ & \best{0.0} \\

        MonST3R (CoTr3) 
        & \third{46.8 / 58.9} & $1.20$ & \third{45.3 / 0.18} & \second{11.7} 
        & $54.7 / 51.5$ & $0.44$ & $44.8 / 0.07$ & \best{0.0} 
        & $56.8 / 52.4$ & $0.70$ & $57.4 / 0.22$ & \best{0.0} \\

        MonST3R (GT depth+CoTr3) 
        & $57.6 / 63.5$ & \third{1.10} & $46.5 / 0.18$ & \second{11.7} 
        & \third{48.6 / 30.1} & $0.41$ & \third{37.6 / 0.06} & \best{0.0} 
        & \third{51.4 / 56.2} & $0.55$ & \third{54.3 / 0.21} & \best{0.0} \\

        SpatialTrackerV2~\cite{xiao2025spatialtracker} 
        & $59.8 / 69.1$ & $1.22$ & $48.6 / 0.19$ & \third{30.0} 
        & $57.9 / 44.6$ & $0.46$ & $41.9 / 0.07$ & \best{0.0} 
        & $59.4 / 49.8$ & \third{0.49} & $57.5 / 0.22$ & \second{20.0} \\

        SpatialTrackerV2 (GT depth) 
        & $48.7 / 69.3$ & \third{1.10} & $46.8 / 0.19$ & \third{30.0} 
        & $54.3 / 38.8$ & $0.43$ & $38.7 / 0.07$ & \best{0.0} 
        & $60.8 / 42.6$ & $0.58$ & $55.6 / 0.22$ & \second{20.0} \\

        BundleSDF~\cite{wen2023bundlesdf} (GT mask) 
        & \second{38.2 / 55.9} & \second{0.95} & \second{23.4 / 0.20} & $55.0$ 
        & \second{26.3 / 24.5} & \second{0.24} & \second{\phantom{1}9.1 / 0.07} & $43.3$ 
        & \second{32.0 / 52.1} & $0.58$ & \second{15.6 / 0.22} & \third{36.7} \\

        ArtGS~\cite{liu2025artgs} 
        & $64.3 / 23.4$ & $1.69$ & $\phantom{1}\text{--} / \text{--}\phantom{1}$ & $95.0$ 
        & $60.5 / 53.2$ & \third{0.35} & $\phantom{1}\text{--} / \text{--}\phantom{1}$ & $66.7$ & $70.1 / 18.2$ & \second{0.34} & $\phantom{1}\text{--} / \text{--}\phantom{1}$ & $66.7$ \\

        \name{} (Ours) 
        & \best{\phantom{1}5.3 / \phantom{1}5.4} & \best{0.03} & \best{\phantom{1}5.0 / 0.02} & \best{\phantom{1}1.7} 
        & \best{12.4 / \phantom{1}1.3} & \best{0.06} & \best{\phantom{1}9.1 / 0.02} & \best{0.0} 
        & \best{\phantom{1}5.6 / \phantom{1}5.5} & \best{0.05} & \best{\phantom{1}8.4 / 0.03} & \best{0.0} \\

        \bottomrule
    \end{tabular}
}
\caption{\textbf{Articulated motion estimation.} 
We report articulation axis direction error ($^\circ$), axis position error (meters; applicable only to revolute joints), joint state error ($^\circ$ for revolute and meters for prismatic joints), and failure rate (\%). 
Values are shown as “XX / YY” for revolute (XX) and prismatic (YY) joints. 
\bestcap{Best}, \secondcap{second-best}, and \thirdcap{third-best} results are highlighted. 
All metrics are lower-is-better.}
\label{tab:articulated_motion_results}
\vspace{-5px}
\end{table*}

\begin{table*}[h]
\centering
\resizebox{\linewidth}{!}{
\begin{tabular}{l*{9}{c}}
\toprule
\multirow{3}{*}{\textbf{Methods}} 
& \multicolumn{3}{c}{\textbf{\syntheticdatasetname{}}} 
& \multicolumn{3}{c}{\textbf{HOI4D}} 
& \multicolumn{3}{c}{\textbf{\realdatasetname{}}} \\
\cmidrule(lr){2-4} \cmidrule(lr){5-7} \cmidrule(lr){8-10}
& \multicolumn{2}{c}{6D part pose} & \multicolumn{1}{c}{Reconstruction}
& \multicolumn{2}{c}{6D part pose} & \multicolumn{1}{c}{Reconstruction}
& \multicolumn{2}{c}{6D part pose} & \multicolumn{1}{c}{Reconstruction} \\
\cmidrule(lr){2-3} \cmidrule(lr){4-4}
\cmidrule(lr){5-6} \cmidrule(lr){7-7}
\cmidrule(lr){8-9} \cmidrule(lr){10-10}
& ADD-S (\%) $\uparrow$ & ADD (\%) $\uparrow$ & CD (cm) $\downarrow$
& ADD-S (\%) $\uparrow$ & ADD (\%) $\uparrow$ & CD (cm) $\downarrow$
& ADD-S (\%) $\uparrow$ & ADD (\%) $\uparrow$ & CD (cm) $\downarrow$ \\
\midrule
MonST3R~\cite{zhang2024monst3r} (ICP) 
& $14.36$ & $\phantom{1}9.62$ & $29.8$ 
& $29.41$ & $19.12$ & $6.4$ 
& $13.21$ & $\phantom{1}9.11$ & $13.4$ \\

MonST3R~\cite{zhang2024monst3r} (CoTr3) 
& \third{33.58} & \third{21.94} & $23.6$ 
& $44.72$ & $25.98$ & $2.1$ 
& $18.64$ & $11.35$ & $17.6$ \\

MonST3R~\cite{zhang2024monst3r} (GT depth+CoTr3) 
& \second{37.12} & \second{30.48} & \third{13.9} 
& $54.83$ & $37.11$ & $1.3$ 
& $20.11$ & \third{14.62} & \third{13.2} \\

SpatialTrackerV2~\cite{xiao2025spatialtracker} 
& $26.94$ & $13.11$ & $34.7$ 
& \second{61.25} & \second{41.02} & \third{0.9} 
& $19.77$ & 13.92 & $18.5$ \\

SpatialTrackerV2~\cite{xiao2025spatialtracker} (GT depth) 
& $29.71$ & $15.43$ & \second{9.88} 
& \third{60.98} & \third{40.21} & \second{0.8} 
& \second{22.94} & \second{15.22} & $13.5$ \\

BundleSDF~\cite{wen2023bundlesdf} (GT mask) 
& $22.84$ & $12.37$ & $17.1$ 
& $53.12$ & $35.22$ & $1.4$ 
& \third{20.88} & $12.91$ & \second{10.6} \\

ArtGS~\cite{liu2025artgs} 
& $\text{--}$ & $\text{--}$ & $20.3$ 
& $\text{--}$ & $\text{--}$ & $6.8$ 
& $\text{--}$ & $\text{--}$ & $20.6$ \\

\name{} (Ours) 
& \best{78.96} & \best{71.28} & \best{\phantom{1}3.2} 
& \best{79.43} & \best{69.85} & \best{0.7} 
& \best{75.62} & \best{68.11} & \best{\phantom{1}6.1} \\

\bottomrule
\end{tabular}
}
\caption{\textbf{6D part pose estimation and reconstruction.} 
We report 6D part pose accuracy using ADD-S and ADD (higher is better) and surface reconstruction quality using Chamfer Distance (CD, lower is better) across the \syntheticdatasetname{}, HOI4D, and \realdatasetname{} datasets. 
}
\vspace{-10px}
\label{tab:part_pose_reconstruction_results}
\end{table*}

\begin{table}[t]
\centering
\scriptsize
\resizebox{\columnwidth}{!}{
\begin{tabular}{l*{3}{c}}
\toprule
\multirow{2}{*}{\textbf{Methods}} & \multicolumn{1}{c}{\textbf{\syntheticdatasetname{}}} & \multicolumn{1}{c}{\textbf{HOI4D}} & \multicolumn{1}{c}{\textbf{\realdatasetname{}}} \\
\cmidrule(lr){2-2} \cmidrule(lr){3-3} \cmidrule(lr){4-4} 
& mIoU $\uparrow$ & mIoU $\uparrow$ & mIoU $\uparrow$ \\
\midrule
MonST3R~\cite{zhang2024monst3r} & \second{23.6} & \second{26.8} & \second{23.7} \\
SpatialTrackerV2~\cite{xiao2025spatialtracker} (SAM2) & \phantom{1}\third{6.2} & \phantom{1}\third{5.8} & \third{13.4} \\
\name{} (Ours) & \best{77.9} & \best{76.4} & \best{74.8}  \\
\bottomrule
\end{tabular}
}
\caption{\textbf{Moving part segmentation.} 
Mean Intersection-over-Union (mIoU) for moving-part segmentation. 
}
\label{tab:part_segmentation_results}
\end{table}

\vspace{-3px}
\paragraph{Datasets.} We evaluate our method on three datasets. The first two, \realdatasetname{} and \syntheticdatasetname{}, are newly collected as described in~\cref{sec:data_collection}. To enable consistent benchmarking with the baselines, we construct a diverse evaluation set of 60 interaction sequences from these datasets, using the annotated interaction intervals. Additionally, we use the HOI4D dataset~\cite{Liu_2022_hoi4d}, which contains short, single-object egocentric RGB-D interaction videos. To adapt it to our setting, we extract 30 interactions involving four articulated object categories (``laptop'', ``cabinet'', ``safe'' and ``trash can'') and post-process the provided 6D part poses to obtain ground-truth articulation parameters.

\paragraph{Evaluation tasks and metrics.}
We consider four tasks:
articulated motion estimation,
6D part pose estimation,
part segmentation, and
3D reconstruction.
For articulated motion estimation, we report axis direction and position errors, and per-timestep joint state error with separate results for revolute and prismatic joints.
Additionally, we include the failure rate, defined as the proportion of videos where the method either failed to process or predicted an incorrect joint type.
For 6D part pose estimation, we follow standard evaluation using ADD-S and ADD metrics~\cite{wen2023bundlesdf, wen2021bundletrack}, while 3D surface reconstruction quality is measured by Chamfer Distance (CD). For segmentation of moving parts we report mean Intersection-over-Union (mIoU).

\paragraph{Baselines.} We compare against three representative categories:
(Type 1) 4D reconstruction pipelines (MonST3R~\cite{zhang2024monst3r}, SpatialTrackerV2~\cite{xiao2025spatialtracker}),
(Type 2) 6D object pose tracking (BundleSDF~\cite{wen2023bundlesdf}), and
(Type 3) articulated object reconstruction (ArtGS~\cite{liu2025artgs}).
For MonST3R, we integrate ICP and CoTracker3~\cite{karaev24cotracker3} to enable moving-part pose tracking. SpatialTrackerV2 is augmented with SAM2 to obtain part segmentations. For both Type 1 baselines, we lift dynamic regions or tracks into 3D and estimate part poses via RANSAC-based model fitting, optionally using ground-truth depth for fair comparison.
BundleSDF is given ground-truth camera poses and segmentation masks to recover canonical part geometry and per-frame part transformations.
ArtGS operates on two static articulation states and cannot track motion in video. Thus, we report its static reconstruction quality and estimated articulation parameters using ground-truth camera poses.
More details are provided in the Supp. Mat.

\paragraph{Articulated motion estimation.}
Results are reported in \cref{tab:articulated_motion_results}, with visual examples in \cref{fig:qualitative_results}.  
\name{} achieves the lowest errors across all datasets and motion types.  
On \syntheticdatasetname{}, it estimates articulation axis direction within $5.3^\circ$ and position within $0.03\,\mathrm{m}$, outperforming BundleSDF~\cite{wen2023bundlesdf} by over $30^\circ$ and an order of magnitude in distance.  
The same trend holds for HOI4D~\cite{Liu_2022_hoi4d} and \realdatasetname{}, where \name{} reduces the state prediction error to $9.1^\circ / 0.02\,\mathrm{m}$ and $8.4^\circ / 0.03\,\mathrm{m}$, respectively, with a failure rate of $0\%$.  
In contrast, existing dynamic reconstruction pipelines~\cite{zhang2024monst3r, xiao2025spatialtracker} and articulated modeling methods~\cite{liu2025artgs} exhibit large deviations and frequent failures.  
These results highlight the reliability of \name{} in recovering physically consistent joint parameters directly from egocentric videos.

\paragraph{Moving part segmentation.}
Results are presented in \cref{tab:part_segmentation_results}.  
Across all datasets, \name{} achieves the highest segmentation accuracy, with mIoU scores of $77.9$ on \syntheticdatasetname{}, $76.4$ on HOI4D, and $74.8$ on \realdatasetname{}.  
These results represent a substantial improvement over prior methods, including MonST3R~\cite{zhang2024monst3r} ($23.6$-$26.8$\,mIoU) and SpatialTrackerV2~\cite{xiao2025spatialtracker} ($5.8$-$13.4$\,mIoU), demonstrating consistently accurate delineation of moving parts across synthetic, controlled, and real-world scenes.

\paragraph{6D part pose estimation.}
\cref{tab:part_pose_reconstruction_results} summarizes 6D part pose accuracy, measured via ADD-S and ADD.  
Across all datasets, \name{} consistently achieves the best results: $79.0\%$ ADD-S / $71.3\%$ ADD on \syntheticdatasetname{}, $79.4\%$ / $69.9\%$ on HOI4D, and $75.6\%$ / $68.1\%$ on \realdatasetname{}.  
These represent more than a twofold improvement over BundleSDF~\cite{wen2023bundlesdf} and a large margin over dynamic trackers~\cite{zhang2024monst3r, xiao2025spatialtracker}.  
The improvements indicate that the proposed joint optimization of part pose and articulation parameters produces stable, temporally consistent motion trajectories.

\paragraph{Articulated reconstruction.}
Finally, \name{} delivers high-fidelity articulated 3D reconstructions, as reported by Chamfer Distance (CD) in \cref{tab:part_pose_reconstruction_results}.  
It achieves $3.2\,\mathrm{cm}$ CD on \syntheticdatasetname{}, $0.7\,\mathrm{cm}$ on HOI4D, and $6.1\,\mathrm{cm}$ on \realdatasetname{}, substantially improving over \textit{all} baselines.  
Examples (\cref{fig:qualitative_results}) show that \name{} reconstructs both static geometry and articulated parts with accurate motion ranges, enabling continuous re-simulation of interactions in 3D.  
Overall, these results confirm that our training-free, optimization-based approach generalizes across synthetic, controlled, and real-world egocentric recordings, producing physically consistent and interactable digital twins.

\vspace{-3px}
\section{Applications}
\label{sec:applications}
\vspace{-3px}

\begin{figure}[h]
    \centering
     \includegraphics[width=\linewidth]{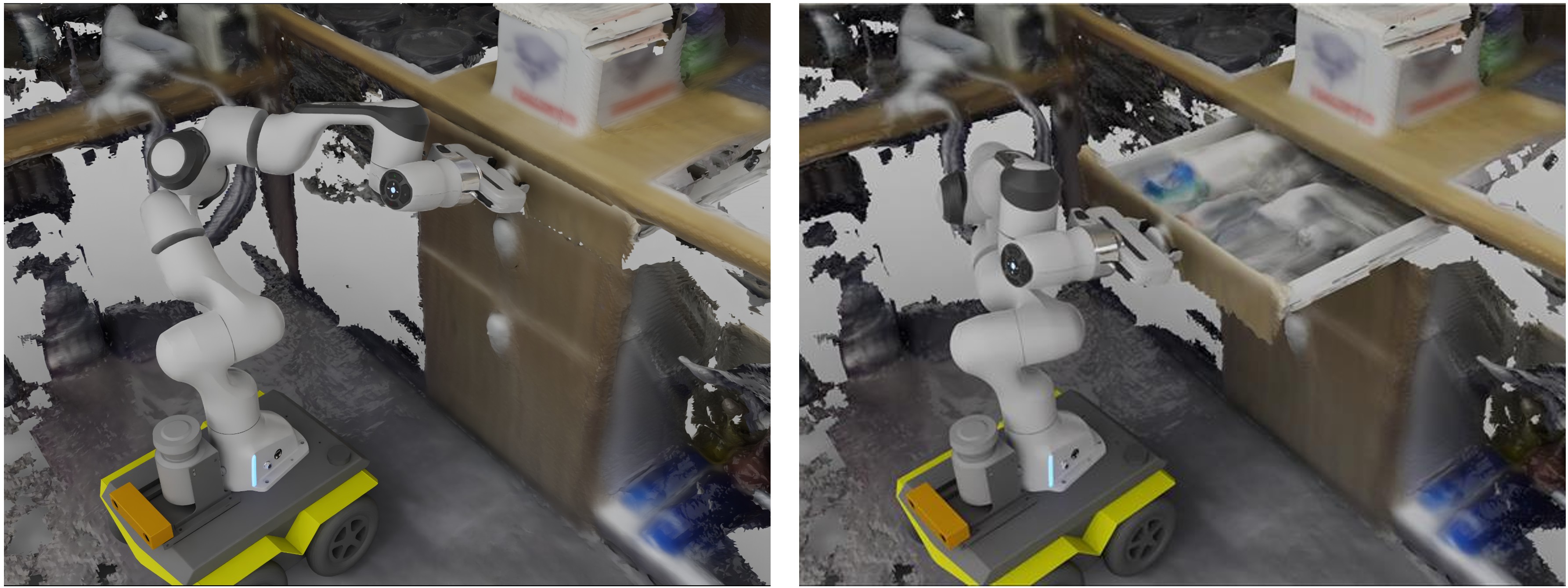}
    \caption{\textbf{Isaac Sim deployment.} A mobile manipulator interacts with a drawer reconstructed from a real-world scan.}
    \label{fig:simulation_format_export}
    \vspace{-2px}
\end{figure}

\paragraph{Simulation-ready export.} Using the reconstructed geometry and articulation information, we generate files that enable interactive digital replicas of real-world scenes for physical simulation. Specifically, we use the articulation parameters to define revolute and prismatic joints between static structures and movable parts, and export them in URDF or USD format. Physical properties such as mass and inertia can be inferred from RGB images by querying a vision-language model (\eg, GPT-5). This process allows the reconstructed scene to be directly loaded into physics simulators, enabling a wide range of downstream tasks such as robot-scene interaction. Fig.~\ref{fig:simulation_format_export} illustrates an example in Isaac Sim~\cite{NVIDIA_Isaac_Sim} where a simulated robot arm interacts with a scene reconstructed by \name{} from real-world scans.

\begin{figure}[h]
\centering
\includegraphics[width=0.45\linewidth,
    trim=0 130px 0 220px,clip]{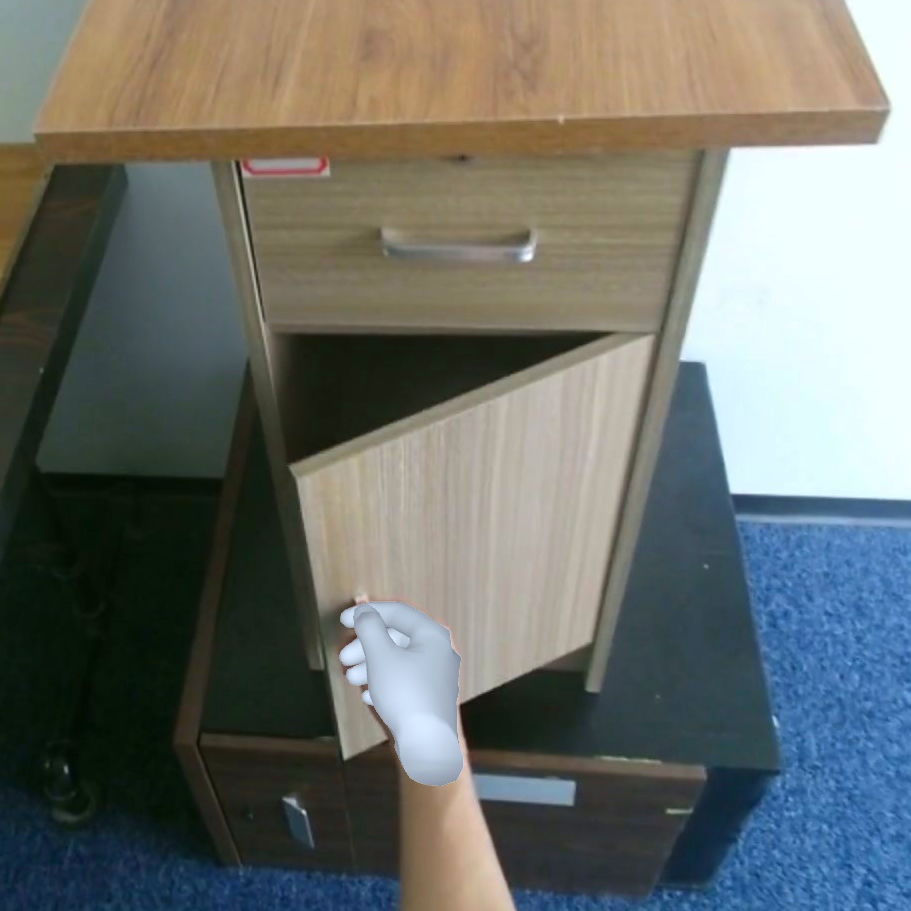}
\includegraphics[width=0.45\linewidth,trim=0 0 0 100
px,clip]{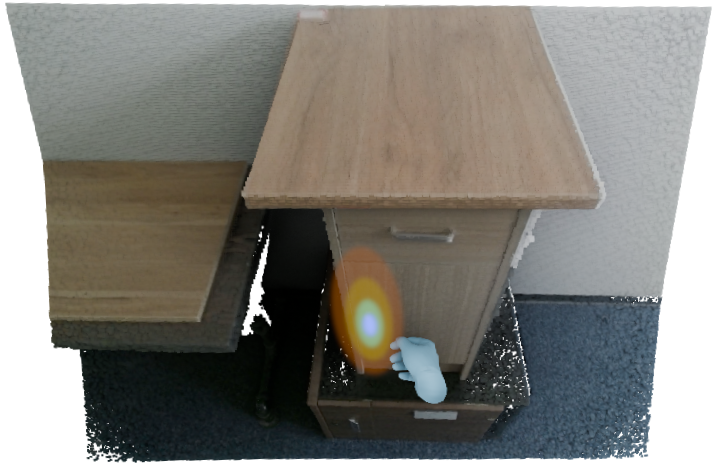}
\caption{\textbf{Hand-scene interaction.} Estimated 3D hand mesh (\emph{left}) and inferred affordance map (\emph{right}). Integrating the hand pose into our functional reconstruction enables finding contact regions and consistent reasoning over the associated scene-part motion.}
\label{fig:hand_object_interaction_application}
\vspace{-1px}
\end{figure}

\paragraph{Hand-guided affordance mapping.} 
As shown in \cref{fig:hand_object_interaction_application}, our framework naturally extends to incorporate hand-centric affordance information. By extracting 3D hand meshes from off-the-shelf estimators~\cite{HaMeR, potamias2024wilor} and aligning them within our functional digital-twin space, we can localize the hand in 3D and recover its contact regions on the object. This enables joint reasoning over the hand’s motion and the motion of the interacted object part.

\begin{figure}[h]
    \centering
     \includegraphics[width=\linewidth]{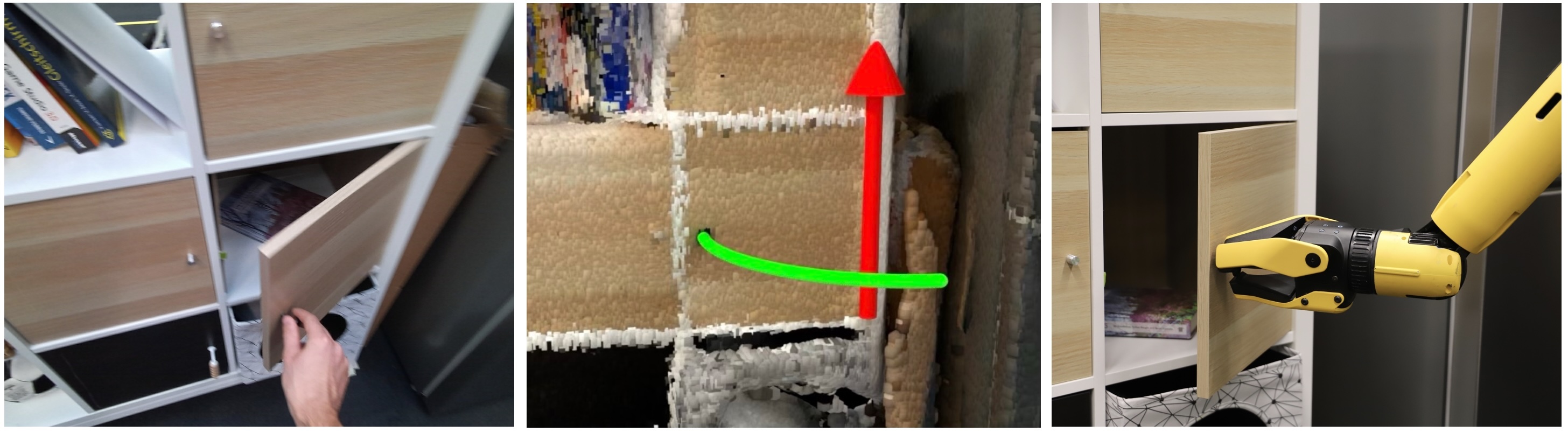}
    \vspace{-15px}
    \caption{\textbf{Robot-scene interaction}.
    \emph{Left:} Human demonstration of opening a cabinet. \emph{Center:} {The articulation trajectory derived from the functional scene model.}
    \emph{Right:} The robot leverages the functional information to reliably reproduce the same interaction.}
    \label{fig:robot_scene_interaction}
\vspace{-2px}
\end{figure}

\paragraph{Robot-scene interaction from human demonstration.} 
The functional scene model can be directly transferred to a mobile manipulator, enabling robot-scene interaction from human demonstrations. \cref{fig:robot_scene_interaction} shows a Boston Dynamics Spot with Arm interacting with articulated objects in the real world. Given the inferred contact points, articulation parameters, and interaction trajectories, the robot can reproduce the same interactions reliably and stably.

\vspace{-4px}
\section{Conclusion}
\label{sec:conclusion}
\vspace{-2px}
We present \name{}, a training-free method to reconstruct functional, articulated 3D digital twins of real environments from a single egocentric RGB-D interaction video. 
By combining geometric reasoning with semantic and motion priors from foundation models, \name{} jointly estimates camera motion, part articulation, and scene geometry. 
Our two new egocentric datasets, \realdatasetname{} and \syntheticdatasetname{}, enable quantitative evaluation and future research on functional scene understanding. 
Experiments across real and simulated settings demonstrate that \name{} substantially outperforms all baselines in articulation estimation, pose estimation, segmentation, and reconstruction quality, and produces digital twins that can be directly used for simulation, affordance reasoning, and robot-scene interaction.

\vspace{4px}
\myparagraph{Acknowledgements} This work was supported by the SNSF Advanced Grant 216260: “Beyond Frozen Worlds: Capturing Functional 3D Digital Twins from the Real World” and the SNSF Postdoc.Mobility grant 222227. The authors also acknowledge the support from a SwissAI Grant for Small Projects and an Academic Grant from NVIDIA. Alexandros Delitzas is also supported by the Max Planck ETH Center for Learning Systems (CLS).

\clearpage
{
    \small
    \bibliographystyle{ieeenat_fullname}
    \bibliography{main}
}


\end{document}